\pdfoutput=1
\documentclass[11pt]{article}
\usepackage{hyperref}
\usepackage[noend]{algpseudocode}
\usepackage{amsmath}
\usepackage{algorithmicx,algorithm}
\usepackage{acl}
\usepackage{float}
\usepackage{times}
\usepackage{latexsym}
\usepackage{graphicx}
\usepackage{bm}
\usepackage[T1]{fontenc}
\usepackage{booktabs}
\usepackage[utf8]{inputenc}

\usepackage{microtype}

\title{Focus on the Target's Vocabulary: \\ Masked Label Smoothing for Machine Translation}

\let\oldtheenumi=\thefootnote

\renewcommand{\thefootnote}{*}

\author{
 Liang Chen, Runxin Xu, Baobao Chang$\footnotemark[1]$
\\ 
 Key Laboratory of Computational Linguistics, Peking University, MOE, China \\
 \texttt{leo.liang.chen@outlook.com} \\
 \texttt{runxinxu@gmail.com}\quad
 \texttt{chbb@pku.edu.cn}
}

\begin{document}
\maketitle
\begin{abstract}
% Label Smoothing and Vocabulary Sharing are two techniques widely applied Machine Translation. However, according to our observation there is a conflict between the two techniques. We analyse the reason for the conflict and propose a plug-and-play module named Masked Label Smoothing to eliminate the problem. Through empirical study, our method shows robust and consistent improvement over original label smoothing on different datasets including bilingual and multilingual translation in both BLEU and calibration scores.

Label smoothing and vocabulary sharing are two widely used techniques in neural machine translation models. However, we argue that simply applying both techniques can be conflicting and even leads to sub-optimal performance. When allocating smoothed probability, original label smoothing treats the source-side words that would never appear in the target language equally to the real target-side words, which could bias the translation model. To address this issue, we propose Masked Label Smoothing (MLS), a new mechanism that masks the soft label probability of source-side words to zero. Simple yet effective, MLS manages to better integrate label smoothing with vocabulary sharing. Our extensive experiments show that MLS consistently yields improvement over original label smoothing on different datasets, including bilingual and multilingual translation from both translation quality and model's calibration. Our code is released at \href{https://github.com/PKUnlp-icler/MLS}{PKUnlp-icler}.

\footnotetext{Corresponding author}
\end{abstract}

\section{Introduction}

Recent advances in Transformer-based~\citep{transformers} models have achieved remarkable success in Neural Machine Translation (NMT). For most NMT studies~\citep{transformers,Song2019MASSMS,RASP,Liu2020MultilingualDP,DBLP:journals/corr/abs-2106-13736}, there are two widely used techniques to improve the quality of the translation:
Label Smoothing (LS) and Vocabulary Sharing (VS).
Label smoothing~\citep{Pereyra2017RegularizingNN} turns the \emph{hard} one-hot labels into a \emph{soft} weighted mixture of the golden label and the uniform distribution over the whole vocabulary, which serves as an effective regularization technique to prevent over-fitting and over-confidence~\citep{Mller2019WhenDL} of the model.
In addition, vocabulary sharing~\citep{Xia2019TiedTN} is another commonly used technique, which unifies the vocabulary of both source and target language into a whole vocabulary, and therefore the vocabulary is shared.
It enhances the semantic correlation between the two languages and reduces the number of total parameters of the embedding matrices.

\begin{figure}[t]
\includegraphics[width=1\columnwidth]{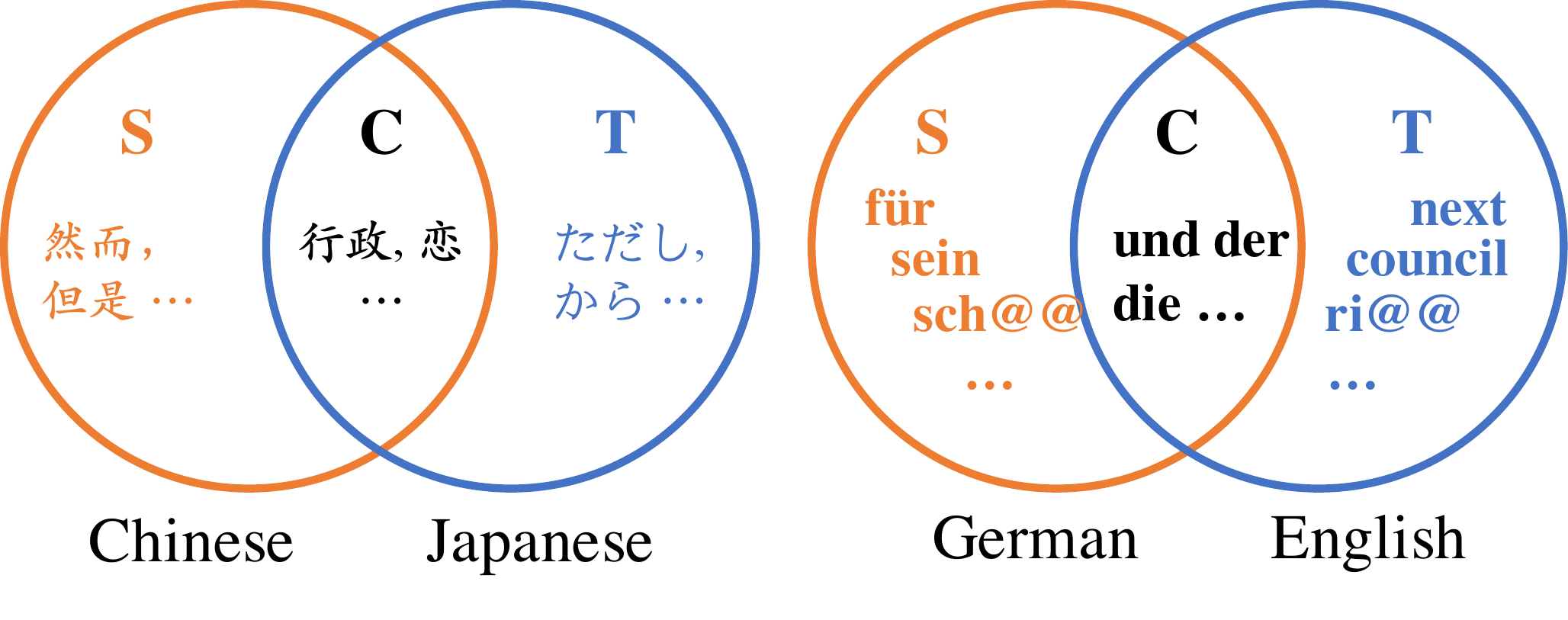}
\caption{Venn diagram showing the structure of the shared vocabulary, which can be divided into three parts: Source (S), Common (C), and Target (T).}
\label{cluste_pic}
\end{figure}

\begin{table}[!t]
\centering
\resizebox{75mm}{13.5mm}{
\begin{tabular}{lcc}
\toprule
\textbf{Model} & \textbf{DE-EN} & \textbf{VI-EN}\\
\midrule
Transformer & 33.54 & 29.95\\
- w/ Label Smoothing (LS)  & \textbf{34.76} & \textbf{30.73} \\ 
- w/ Vocabulary Sharing (VS) & 33.83 & 29.36\\
- w/ LS+VS $^\dagger$ & 34.56 & 30.41\\
\bottomrule
\end{tabular}}
\caption{Results in IWSLT'14 DE-EN and IWSLT'15 VI-EN datasets.$\dagger$ denotes consistent setting to \citet{transformers}. Jointly adopting label smoothing and vocabulary sharing techniques cannot achieve further improvements, but leads to sub-optimal performance.}
\label{tab:pre_study}
\end{table}

However, in this paper, we argue that jointly adopting both label smoothing and vocabulary sharing techniques can be conflicting, and leads to sub-optimal performance.
Specifically, with vocabulary sharing, the shared vocabulary can be divided into three parts as shown in Figure~\ref{cluste_pic}.
But with label smoothing, the soft label still considers the  words at the source side that are impossible to appear at the target side.
This would mislead the translation model and exerts a negative effect on the translation performance.
As shown in Table~\ref{tab:pre_study}, although introducing label smoothing or vocabulary sharing alone can improve the vanilla Transformer, jointly adopting both of them cannot obtain further improvements but achieves sub-optimal results.

% We argue that the uniform distribution over the shared vocabulary space is to blame for the conflict that it misleads the translation system to allocate redundant probability to those "impossible" labels when executing label smoothing. 

% In this paper we analyse the reason for the conflict and propose a plug-and-play module named Masked Label Smoothing (MLS) to solve the problem. The mask can be generated with the source, target and joint dictionary via simple set operations like union and intersect. MLS on the one hand, keeps the benefits of Label Smoothing and Vocabulary Sharing. On the other hand, eliminates the conflict of the two techniques.

To address the conflict of label smoothing and vocabulary sharing, we first propose a new mechanism named Weighted Label Smoothing (WLS) to control the smoothed probability distribution and its parameter-free version Masked Label Smoothing (MLS). 
Simple yet effective, MLS constrains the soft label not to assign soft probability to the words only belonging to the source side.
In this way, we not only keeps the benefits of both label smoothing and vocabulary sharing, but also address the conflict of these two techniques to improve the quality of the translation.
% The mask can be generated with the source, target and joint dictionary via simple set operations like union and intersect. MLS on the one hand, keeps the benefits of Label Smoothing and Vocabulary Sharing. On the other hand, eliminates the conflict of the two techniques.

% The idea of MLS is intuitive and originates from real-world scenarios. When native speakers are doing translation, they tend to focus and only select word or phrases which come from the target language's vocabulary. We formulate the knowledge by looking up the source, target, joint vocabulary as depicted in Figure~\ref{cluste_pic} and generate a scale mask. During training we inject the external knowledge to the model by smoothing different labels at different scales according to the scale mask.

According to our experiments, MLS leads to a better translation not only in scores like BLEU but also reports improvement in model's calibration. Compared with original label smoothing with vocabulary sharing, MLS outperforms in WMT'14 EN-DE(+0.47 BLEU), WMT'16 EN-RO (+0.33 BLEU) and other 7 language pairs including DE,RO-EN multilingual translation task. 

\section{Background}

\paragraph{Label Smoothing}
The original label smoothing can be formalized as:
\begin{equation}
\bm{\hat{y}}^{LS} = \bm{\hat{y}}(1-\alpha) + \bm{\alpha}/K
\end{equation}

$K$ denotes the number of classes, $\alpha$ is the label smoothing parameter, $\bm{\alpha}/K$ is the soft label, $\bm{\hat{y}}$ is a vector where the correct label equals to 1 and others equal to zero and $\bm{\hat{y}}^{LS}$ is the modified targets.

Label smoothing is first introduced to image classification~\citep{Szegedy2016RethinkingTI} task. 
\citet{Pereyra2017RegularizingNN,Edunov2018ClassicalSP} explore label smoothing's application in Sequence generation from token level and \citet{NIPS2016_seq_ls} propose sentence level's label smoothing. Theoretically, \citet{Mller2019WhenDL,meister-etal-2020-generalized} all point out the relation between label smoothing and entropy regularization. \citet{gao-etal-2020-towards} explores the best recipe when applying label smoothing to machine translation. To generate more reliable soft labels, \citet{Lukasik2020SemanticLS} takes semantically similar n-grams overlap into consideration level label smoothing. \citet{wang-etal-2020-inference} proposes Graduate Label Smoothing that generate soft label according to the different confidence scores of model.  To the best of our knowledge, we are the first to investigate label smoothing's influence on machine translation from the perspective of languages.

\paragraph{Vocabulary Sharing} Vocabulary sharing is widely applied in most neural machine translation studies~\citep{transformers,Song2019MASSMS,RASP}. Researchers have conducted in-depth studies in Vocabulary Sharing. \citet{Liu2019SharedPrivateBW} propose shared-private bilingual word embeddings, which give a closer relationship between the source and target embeddings. While \citet{kim-etal-2019-effective} point out that there is an vocabulary mismatch between parent and child languages in shared multilingual word embedding.
\section{Conflict Between Label Smoothing and Vocabulary Sharing}

Words or subwords in a language pair's joint dictionary can be categorized into three classes: \textbf{source}, \textbf{common} and \textbf{target} using Venn Diagram according to their belonging to certain language as depicted in Figure~ \ref{cluste_pic}. This can be achieved by checking whether one token in the joint vocabulary also belongs to the source/target vocabulary. We formalized the categorization algorithm in Appendix~\ref{sec:appendix_alg}. 

Then we compute the tokens' distribution in different translation directions as shown in Table~\ref{tab:token_distribution}. Tokens in source class account for a large proportion up to 50\%. When label smoothing and vocabulary sharing are together applied, the smoothed probability will be allocated to words that belong to the source class. Those words have zero overlap with the possible target words, therefore they have no chance to appear in the target sentence. Allocating smoothed probability to them might introduce extra bias for the translation system during training process, unavoidably leading to a higher translation perplexity as also revealed by \citet{Mller2019WhenDL}.

\begin{table}[t]
\centering

\begin{tabular}{lccc}
\toprule
\textbf{Category} & \textbf{DE->EN} & \textbf{RO->EN} & \textbf{VI->EN}\\\midrule
Source & 39\% & 50\%& 36\% \\
Common & 20\% & 8\% & 11\% \\
Target & 41\% & 42\% & 53\% \\

\bottomrule
\end{tabular}
\caption{The distribution of different categories of the shared vocabulary forWMT'14 DE-EN, WMT'16 RO-EN, and IWSLT'15 VI-EN datasets. The proportion of tokens belonging to source category is up to 50\%, which might mislead the translation model.}
\label{tab:token_distribution}
\end{table}

Table~\ref{bilingual_tab} reveals the existence of conflict, that the joint use of label smoothing and vocabulary sharing doesn't compare with solely use one technique in all language pairs with a maximum loss of 0.32 BLEU score. 

%Label Smoothing with uniform distribution is consistent to the theory of max entropy that if no extra knowledge is given, we are recommended to choose the model with maximum entropy. 

% In practice of original Label Smoothing, we give all incorrect tokens one same smoothed probability to achieve maximum entropy. However, under the setting of Vocabulary Sharing, we actually have such strong prior information that tokens don't belong to the target side shouldn't be allocated as much probability as tokens which are really shared by two languages.

% \citet{Li2020DatadependentGP} utilizes the semantic similarity by introducing a continuous Gaussian prior computed with word vectors to the objective function and proves the effectiveness. In terms of semantic correlations, those "source only" tokens barely have semantic similarity to the correct token since they seldom or never co-exist in one sentence. Unlike \citet{Li2020DatadependentGP}'s continuous prior, our prior information is a kind of truncation that calls for masking mechanism to inject the knowledge to the model.

\section{Methods}

\def\@fnsymbol#1{\ensuremath{\ifcase#1\or *\or \dagger\or \ddagger\or
   \mathsection\or \mathparagraph\or \|\or **\or \dagger\dagger
   \or \ddagger\ddagger \else\@ctrerr\fi}}
   
\begin{table*}[!t]
    \centering
    (a) Bilingual Translation
    
    \resizebox{140mm}{18.5mm}{
    
    \begin{tabular}{l|c|c|c|c|c|c|c}
    \toprule
         &  \multicolumn{2}{c|}{\textbf{WMT'16}}  & \textbf{IWSLT'14} &\multicolumn{2}{c|}{\textbf{WMT'14}}  & \textbf{IWSLT'15}& \textbf{CASIA} \\ \midrule
        \textbf{Model} & RO-EN & EN-RO & DE-EN & DE-EN & EN-DE & VI-EN & ZH-EN \\ 
        \midrule
        
        Transformer & 22.03 &  19.61 &  33.54 & 30.85 & 27.21 & 29.95 & 20.66 \\

        - w/ VS  & 22.20 &  19.91 &  33.83 & 31.08 & 27.51 & 29.36 & 20.88 \\
        - w/ LS & 22.96 &  20.68 & 34.76 & 31.14 & 27.53 & \textbf{30.73} & 21.10 \\

        - w/ LS+VS   & 22.89 & 20.59 & 34.56 & 30.98 &  27.44 & 30.41 & 21.04 \\ 

        -  w/ MLS (ours) & \textbf{23.22**} & \textbf{20.88**} &\textbf{35.04**} & \textbf{31.43*} & \textbf{27.91*} & 30.57* & \textbf{21.23*} \\
        \bottomrule 
    \end{tabular}}
    
    \vspace{1.5ex} 
    
    (b) Multilingual Translation 
    \resizebox{120mm}{11.5mm}{
    \begin{tabular}{l|c|c|c|c|c|c}
\toprule
& \multicolumn{3}{c|}{\textbf{IWSLT'14+WMT'16}} & \multicolumn{3}{c}{\textbf{IWSLT'14+WMT'16$\dagger$}}\\  
\midrule 
\textbf{Model}& \textbf{DE,RO-EN} & \textbf{DE-EN} & \textbf{RO-EN}&\textbf{DE,RO-EN} & \textbf{DE-EN} & \textbf{RO-EN}\\
\midrule
 - w/ LS+VS& 33.78 & 37.24 & 23.15 & 33.25 & 37.44 & 20.40 \\
 - w/ MLS (ours) & \textbf{34.10**} &  \textbf{37.53**} & \textbf{23.19} & \textbf{33.53**} & \textbf{37.77**} & \textbf{20.86**}\\
 \bottomrule
\end{tabular}}

    \caption{Results of bilingual translation tasks (a) and multilingual translation (b).  $\dagger$ denotes the balanced version of multilingual translation data. Same conflict between LS and VS occurs in all language pairs. Our MLS outperforms the original label smoothing with vocabulary sharing with significance levels when of p < 0.01 (**), p < 0.05 (*) and also beats individually using LS or VS in most cases.}
\label{bilingual_tab}
\end{table*}

\subsection{Weighted Label Smoothing}

To deal with the conflict when executing label smoothing, we propose a plug-and-play Weighted Label Smoothing mechanism to control the smoothed probability's distribution.  

Weighted Label Smoothing(WLS) has three parameters $\beta_{t}$, $\beta_{c}$, $\beta_{s}$ apart from the label smoothing parameter $\alpha$, where the ratio of the three parameters represents the portion of smoothed probability allocated to the target, common and source class and the sum of the three parameters is 1. The distribution within token class follows a uniform distribution. WLS can be formalized as: 
\begin{equation}
\bm{\hat{y}}^{WLS} = \bm{\hat{y}}(1-\alpha) + \bm{\beta} \\
\end{equation}

where $\bm{\hat{y}}$ is a vector where the element corresponding to the correct token equals to 1 and others equal to zero. $\bm{\beta}$ is a vector that controls the distribution of probability allocated to incorrect tokens. We use $t_i,c_i,s_i$ to represent probability allocated to the i-th token in the target,common,source category, all of which form the distribution controlling vector $\bm{\beta}$ with $\sum_i^{K}{\beta_i} = \alpha$. The restriction can be formalized as:
\begin{equation}
\sum{t_i}:\sum{c_i}:\sum{s_i} = \beta_{t}:\beta_{c}: \beta_{s}
\end{equation}

\subsection{Masked Label Smoothing}

Based on the Weight Label Smoothing mechanism, we can now implement Masked Label Smoothing by set $\beta_s$ to 0 and regard the target and common category as one category. In this way, Masked Label Smoothing is parameter-free and implicitly injects external knowledge to the model. And we have found out that this simple setting can reach satisfactory results according our experiments.

\begin{figure}[!h]
\centering
\includegraphics[width=1.0\columnwidth]{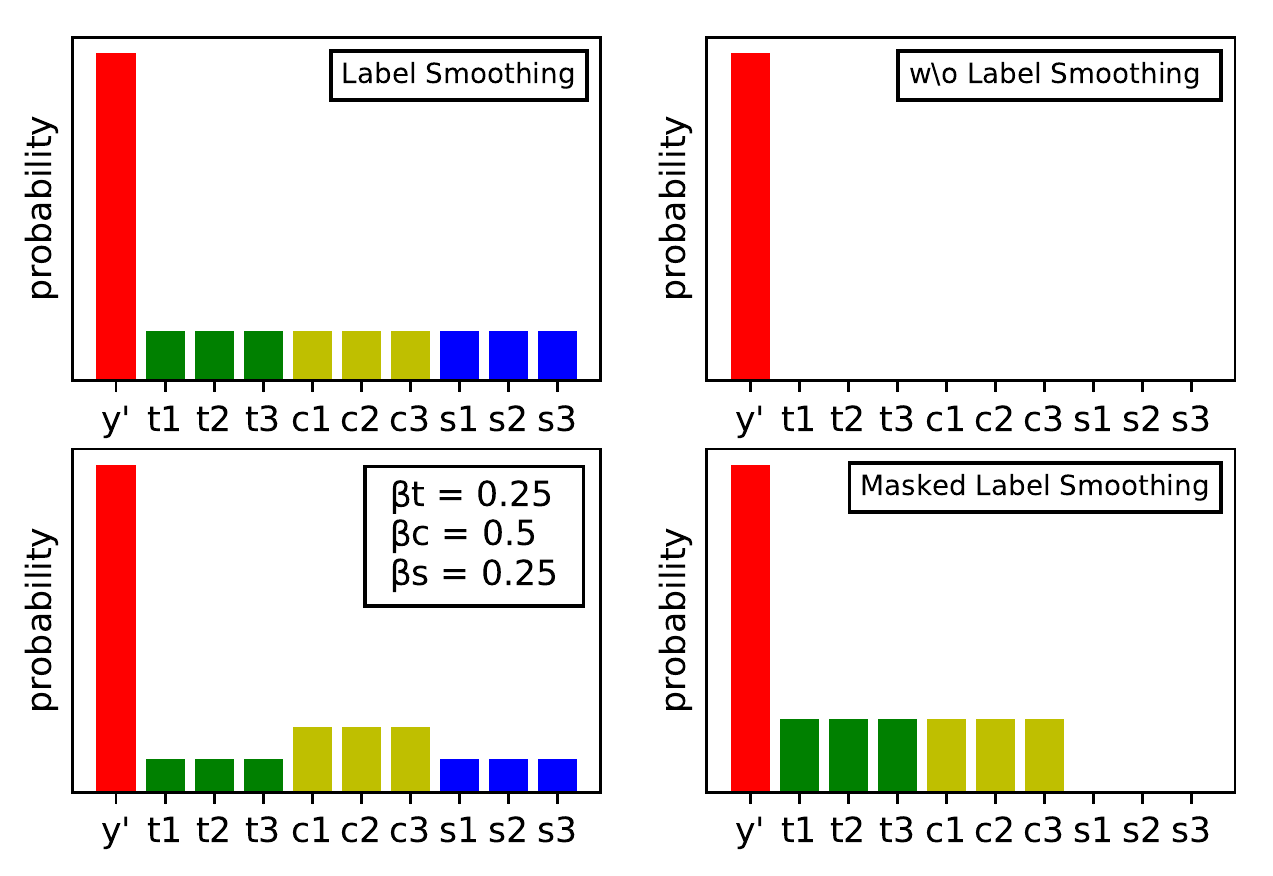}
\caption{Illustration of different label smoothing methods. The height of each bar in the graph denoted the probability allocated to each token. $y'$ is the current token during current decoding phase. We assume that there are only 10 tokens in the joint vocabulary and t1-t3 belongs to target class, c1-c3 belongs to common class and s1-s3 belongs to source class.}
\label{bar}
\end{figure}

We illustrate different label smoothing methods in Figure~\ref{bar}. It is worth noticing that MLS is different from setting WLS's parameters to 1-1-0 since there might be different number of tokens in the common and target vocab.

\section{Experiments}

\subsection{Task Settings}

For bilingual translation, we conduct experiments on 7 translation tasks. We choose language pairs that have different ratio of common sub-words. These include WMT'14 DE-EN,EN-DE, IWSLT'14 DE-EN, IWSLT'15 VI-EN, WMT'16 RO-EN,EN-RO and CASIA ZH-EN. 

We use the official train-dev-test split of WMT'14, 16 and IWSLT'14, 15 datasets. For CASIA ZH-EN dataset, we randomly select 5000 sentences as development set and 5000 sentences as test set from the total dataset. 

For multilingual translation, we combine the WMT'16 RO-EN and IWSLT'14 DE-EN datasets to formulate a RO,DE-EN translation task. We also make a balanced multilingual dataset that has equal numbers of DE-EN and RO-EN training examples to reduce the impact of imbalance languages and to explore how MLS performs under different data distribution condition in multilingual translation.

We apply the Transformer
base ~\citep{transformers} model as our baseline model. We fix the label smoothing parameter $\alpha$ to 0.1 in the main experiments and individually experiment and examine the performance of MLS under different $\alpha$. 

\renewcommand{\thefootnote}{\oldtheenumi}

We use compound\_split\_bleu.sh from fairseq to compute the final bleu scores. The inference ECE score\footnote{https://github.com/shuo-git/InfECE} and chrF score\footnote{https://github.com/m-popovic/chrF} are computed through open source scripts. We list the concrete training and evaluation settings in Appendix~\ref{sec:appendx_exp}.

% \begin{table}[!t]
% \centering
% \begin{tabular}{l|c|c|c}
% \hline
% & \multicolumn{3}{c}{\textbf{IWSLT'14+WMT'16*}} &
% \hline
% \textbf{Model}& \textbf{DE,RO-EN} & \textbf{DE-EN} & \textbf{RO-EN}\\
% \hline
%  \textbf{w/ LS+VS}& 33.25 & 37.44 & 20.40 \\
%  \textbf{w/ MLS} & \textbf{33.53} & \textbf{37.77} & \textbf{20.86}\\
% \end{tabular}
% \caption{Result of DE,RO-EN Balanced Multilingual Translation, MLS surpass the original LS in both DE-EN,RO-EN directions}
% \label{tab:multi_balance}
% \end{table}

% For multilingual translation, we combine the training, dev and test set of IWSLT'14 DE-EN and WMT'16 RO-EN. In the balanced version, we choose the first 160239 sentence pairs in the WMT'16 RO-EN training set, which is the same as the total number of examples in IWSLT'14 DE-EN training set. Due to the space limitation, the full settings is listed in Appendix~\ref{sec:appendx_exp}.

\subsection{Results}

\paragraph{Bilingual}
Table ~\ref{bilingual_tab} shows the results of bilingual translation experiments. The results reveal the conflict between LS and VS that models with only LS surpass models with both LS and VS in all experiments. Our Masked Label Smoothing obtained consistent improvements over original LS+VS in all tested language pairs significantly.

The effectiveness of MLS maintained under different $\alpha$ value as shown in Table~\ref{tab:1} for both BLEU and chrF scores. Similar to \citet{gao-etal-2020-towards}'s conclusion, we find that a higher $\alpha$ can generally improve the bilingual translation quality. And applying MLS can further improve the results. It shows that not only the probability increase in target vocabulary, but also the allocation of smoothed probabilities in different languages matters in the improvement of translation performance.

\paragraph{Multilingual}
As shown in Table~\ref{bilingual_tab}, MLS achieves consistent improvement over the original label smoothing in both the original and the balanced multilingual translation dataset under all translation directions. In the original combined dataset, direction RO-EN (400K) has much more samples than DE-EN (160K). We do not apply a resampling strategy during training in order to investigate how the imbalance condition affects different models' performance. The balanced version cuts down samples in RO-EN direction to the same number as in DE-EN direction. 

Compared with the imbalance version, the balanced version gave better BLEU scores in DE-EN direction while much worse performance in RO-EN translation for both the original label smoothing and MLS. It indicates that the cut down on RO-EN training examples does weaken the generalization of model in RO-EN translation however doesn't influence the DE-EN translation quality since the RO-EN data might introduce bias to the training process for DE-EN translation.

\begin{table}[t]
\centering
    (a) EN-RO
  \resizebox{0.49\textwidth}{!}{%
  
\begin{tabular}{lcrr}
\toprule
\multicolumn{1}{l|}{Scores} & \multicolumn{3}{c}{BLEU(chrF)}                                                      \\ \midrule
\multicolumn{1}{c|}{ $\alpha$}         & 0.1                                & \multicolumn{1}{c}{0.3} & \multicolumn{1}{c}{0.5}        \\
\multicolumn{1}{l|}{LS+VS}         & \multicolumn{1}{r}{20.54(45.54)}          & 20.65(45.79)                   & \multicolumn{1}{r}{20.62(45.7)}                         \\
\multicolumn{1}{l|}{MLS}           & \multicolumn{1}{r}{\textbf{20.57(45.68)}} & \textbf{20.99(46.29)}          & \multicolumn{1}{r}{\textbf{21.10(46.4)}} \\
\bottomrule

\end{tabular}}

    \vspace{2ex} 
    (b) RO-EN
  \resizebox{0.49\textwidth}{!}{%
\begin{tabular}{lcrr}

\toprule

\multicolumn{1}{l|}{Scores} & \multicolumn{3}{c}{BLEU(chrF)}                                                      \\ \midrule
\multicolumn{1}{c|}{ $\alpha$}         & 0.1                                & \multicolumn{1}{c}{0.3} & \multicolumn{1}{c}{0.5}          \\
\multicolumn{1}{l|}{LS+VS}         & \multicolumn{1}{r}{22.54(47.09)}          & 22.95(47.29)                   & \multicolumn{1}{r}{22.98(47.23)}                   \\
\multicolumn{1}{l|}{MLS}           & \multicolumn{1}{r}{\textbf{22.89(48.23)}} & \textbf{23.10(48.36)}          & \multicolumn{1}{r}{\textbf{23.07(47.39)}}        \\
\bottomrule         
\end{tabular}}

\caption{Individual experiment on $\alpha$. BLEU and chrF scores are reported under different label smoothing  $\alpha$ on WMT'16 EN-RO (a) and RO-EN (b) datasets.}
\label{tab:1}
\end{table}

Even under imbalance condition, MLS can give a better performance (37.53) compared to original LS in the balance condition (37.44). It implies that MLS can relieve the imbalance data issue in multilingual translation. However, the improvement in relative high-resources direction (RO-EN) is not as significant as in the balanced condition. We guess that label smoothing has more complex influence on multilingual model due to the increase of languages and relation among different languages. We leave those questions for future exploration.

% \begin{table*}[!t]
% \centering
% \begin{tabular}{l|c|c|c|c|c|c}
% \hline
% & \multicolumn{3}{c|}{\textbf{IWSLT'14+WMT'16}} & \multicolumn{3}{c}{\textbf{IWSLT'14+WMT'16*}} & 
% \hline
% \textbf{Model}& \textbf{DE,RO-EN} & \textbf{DE-EN} & \textbf{RO-EN}&\textbf{DE,RO-EN} & \textbf{DE-EN} & \textbf{RO-EN}\\
% \hline
%  \textbf{w/ LS+VS}& 33.78 & 37.24 & 23.15 & 33.25 & 37.44 & 20.40 \\
%  \textbf{w/ MLS} & \textbf{34.10} &  \textbf{37.53} & \textbf{23.19} & \textbf{33.53} & \textbf{37.77} & \textbf{20.86}\\
% \end{tabular}
% \caption{Result of DE,RO-EN Multilingual Translation, MLS surpass the original LS in both DE-EN,RO-EN directions. $*$ denotes the balanced version of multilingual translation data.}
% \label{tab:multi_study}
% \end{table*}

\section{Discussion}

\begin{table}[t]
    \centering

    \begin{tabular}{c|c|c|c|c|c}
    \toprule
        $\bm{\beta_{t}}$ & $\bm{\beta_{c}}$  & $\bm{\beta_{s}}$ & \textbf{RO-EN}& \textbf{EN-RO}& \textbf{DE-EN}   \\ \midrule
        - & - & - & 22.80 & 23.15 & 30.94\\ 
        1/3 & 1/3 & 1/3 & 22.68 & 23.19 & \textbf{31.40}\\ 
        1/2 & 1/2 & 0 & \textbf{23.05} & 23.19 & 31.18 \\ 
        1/2 & 0 & 1/2 & 22.86 & 23.01 & 31.33\\ 
        0 & 1/2 & 1/2 & 22.22 & \textbf{23.33} & 30.85 \\ 
        1/2 & 1/4 & 1/4 & 22.73 & 23.16 & 30.92\\ \bottomrule
    \end{tabular}
    \caption{Value "-" denotes the original label smoothing. WLS generally can improve the translation quality with appropriate parameters. Scores are computed using the development set of each direction.}
    \label{ablation}
\end{table}

\subsection{Exploring of Weighted Label Smoothing}

As reported in Table~\ref{ablation}, we explore the influence of different WLS on multiple tasks including WMT'16 RO-EN,EN-RO and WMT'14 DE-EN. 

According to the result, though the best BLEU score's WLS setting vary from different tasks and there seems to exist a more complex relation between the probability allocation and the BLEU score, we still have two observations. First, applying WLS can generally boost the quality of translation compared to the original label smoothing. Second, only WLS with $\beta_{t}$, $\beta_{c}$, $\beta_{s}$ each equals to 1/2-1/2-0 can outperform the original label smoothing on all tasks, which suggests the setting is the most robust one. Thus we recommend using this setting as the initial setting when applying WLS. 

Furthermore, the most robust setting agrees with the form of MLS since they both allocate zero probability to the source category's tokens, which further proves the robustness of MLS.

\subsection{Improvement in Model's Calibration and Translation Perplexity}

\citet{Mller2019WhenDL} have pointed out label smoothing prevents the model from becoming over-confident therefore improve the calibration of model. Since there is a training-inference discrepancy in NMT models, inference ECE score \citep{wang-etal-2020-inference} better reflects models' real calibration.

To compute the ECE scores, we need to split the model's predictions into $M$ bins according to the output confidence and calculate the weighted average of bin's confidence/accuracy difference as the ECE scores considering the number of samples in each bin. 
$$
ECE=\sum_{i=1}^{M} \frac{\left|B_{i}\right|}{N}\left|\operatorname{acc}\left(B_{i}\right)-\operatorname{confidence}\left(B_{i}\right)\right|
$$

where $N$ is the number of total prediction samples and
$B_{i}$ is the number of samples in the $i$-th bin. $\operatorname{acc}\left(B_{i}\right)$ is the average accuracy in the $i$-th bin.

The score denotes the difference between accuracy and confidence of models' output during inference. Less ECE implies better calibration.

The inference ECE scores of our models are shown in Table~\ref{tab:cal_study}. It turns out that models with MLS have lower Inference ECE scores on different datasets. The results indicate that MLS will lead to better model calibration.

We also find out that MLS leads to a significantly lower perplexity than LS during the early stage of training in all of our experiments. It's not surprising since zeroing the source side words' smoothed probability can decrease the perplexity. It can be another reason for model's better translation performance since it gives a better training initialization.

% \begin{table*}[t]
%     \centering
%     \begin{tabular}{l|c|c|c|c|c|c|c|c}
%     \hline
%          & \multicolumn{2}{c|}{\textbf{DE-EN}}& \multicolumn{2}{c|}{\textbf{VI-EN}}  &\multicolumn{2}{c|}{\textbf{DE,RO-EN}}  & \multicolumn{2}{c} {\textbf{DE,RO-EN^{*}}} & \hline
%         \textbf{Model} & BLEU & ECE & BLEU & ECE & BLEU & ECE & BLEU & ECE \\ \hline
%         Transformer & 30.98& 9.77 & 30.40 & 13.07& 33.78 & 11.62  & 33.25 &  10.77\\ 
%         Transformer\textbf{+MLS} & \textbf{31.43}& \textbf{9.67} & \textbf{30.54} & \textbf{12.63} & \textbf{34.10} & \textbf{11.37} & \textbf{33.53} & \textbf{8.82} \\ \hline
%     \end{tabular}
% \caption{BLEU4 and inference ECE score on different translation tasks. $*$ denotes the balanced version of multilingual translation data.}
% \label{tab:cal_study}
% \end{table*}

\begin{table}[t]
\centering
\resizebox{77.5mm}{8mm}{
\begin{tabular}{lcccc}
\toprule
\textbf{Model} & \textbf{DE-EN} & \textbf{VI-EN} & \textbf{DE,RO-EN}& \textbf{DE,RO-EN*} \\\midrule
 - w/ LS+VS& 9.77 & 13.07& 11.62 &  10.77 \\
 - w/ MLS & \textbf{9.67}  & \textbf{12.63} &  \textbf{11.37} &  \textbf{8.82} \\
\bottomrule
\end{tabular}}
\caption{Inference ECE score (less is better) on different translation tasks. $*$ denotes the balanced version of multilingual data. MLS leads to an average of 0.7 lower ECE score, suggesting better model calibration.}
\label{tab:cal_study}
\end{table}

\section{Conclusion}

We reveal the conflict between label smoothing and vocabulary sharing techniques in NMT that jointly adopting the two techniques can lead to sub-optimal performance.
To address this issue, we introduce Masked Label Smoothing to eliminate the conflict by reallocating the smoothed probabilities according to the languages' differences. 
Simple yet effective, MLS shows improvement over original label smoothing from both translation quality and model's calibration on a wide range of tasks.

\section{Acknowledgements}
We thank all reviewers for their valuable suggestions for this work. This paper is supported by the National Science Foundation of China under Grant No.61876004 and 61936012, the National Key Research and Development Program of China under Grant No. 2020AAA0106700.

\section{Ethics Consideration}
We collect our data from public datasets that permit academic use. The open-source tools we use for training and evaluation are freely accessible online without copyright conflicts.

% This document has been adapted \cite{nallapati2016summarunner}
% by Steven Bethard, Ryan Cotterell and Rui Yan
% from the instructions for earlier ACL and NAACL proceedings, including those for 
% ACL 2019 by Douwe Kiela and Ivan Vuli\'{c},
% NAACL 2019 by Stephanie Lukin and Alla Roskovskaya, 
% ACL 2018 by Shay Cohen, Kevin Gimpel, and Wei Lu, 
% NAACL 2018 by Margaret Mitchell and Stephanie Lukin,
% Bib\TeX{} suggestions for (NA)ACL 2017/2018 from Jason Eisner,
% ACL 2017 by Dan Gildea and Min-Yen Kan, 
% NAACL 2017 by Margaret Mitchell, 
% ACL 2012 by Maggie Li and Michael White, 
% ACL 2010 by Jing-Shin Chang and Philipp Koehn, 
% ACL 2008 by Johanna D. Moore, Simone Teufel, James Allan, and Sadaoki Furui, 
% ACL 2005 by Hwee Tou Ng and Kemal Oflazer, 
% ACL 2002 by Eugene Charniak and Dekang Lin, 
% and earlier ACL and EACL formats written by several people, including
% John Chen, Henry S. Thompson and Donald Walker.
% Additional elements were taken from the formatting instructions of the \emph{International Joint Conference on Artificial Intelligence} and the \emph{Conference on Computer Vision and Pattern Recognition}.

% % Entries for the entire Anthology, followed by custom entries

\bibliography{anthology,custom}
\bibliographystyle{acl_natbib}

\newpage
\appendix

\section{Algorithm}
\label{sec:appendix_alg}
\begin{algorithm}[h]
\caption{Divide Token Categories} %算法的名字
\hspace*{0.02in} {\bf Input:} %算法的输入， \hspace*{0.02in}用来控制位置，同时利用 \\ 进行换行
List: S, T, J\\
\hspace*{0.02in} {\bf Output:} %算法的结果输出
List: A,B,C \\
\hspace*{0.02in} {\bf Description:} S is the vocabulary list for source language, T for target language, J for joint vocabulary. A is the output vocabulary for source tokens, B for common tokens, C for target tokens.
\begin{algorithmic}[1]
\State Initialize empty list A,B,C
\For{i in J} % For 语句，需要和EndFor对应
\If{i in S and i in T}
\State B.add(i)
\Else
\If{i in S}
\State A.add(i)
\Else
\State C.add(i)
\EndIf
\EndIf
\EndFor
\State \Return A,B,C
\end{algorithmic}
\end{algorithm}

% \section{Formulation of Label Smoothing with Scale Mask}
% \label{sec:appendix}
% Label Smoothing with scale mask can be formalized as following equations:

% \begin{equation}
% \bm{\hat{y}}^{MLS} = \bm{\hat{y}}(1-\alpha) + \bm{t}+ \bm{c} +\bm{s} \\
% \end{equation}

% where $\bm{\hat{y}}$ is a vector where the element corresponding to the correct token equals to 1 and others equal to zero. $\bm{t},\bm{c},\bm{s}$ are vectors representing smoothed probabilities allocated to the target, common and source tokens. The ratio and quantity of the allocation are controlled by parameter $\alpha,\beta,\gamma,\epsilon$ as formalized in following equations:

% \begin{equation}
%     \sum{t_i}:\sum{c_i}:\sum{s_i} = \beta:\gamma:\epsilon \\  
% \end{equation}

% \begin{equation}
%     \sum{t_i}+\sum{c_i}+\sum{s_i} = \alpha
% \end{equation}

\section{Experiment Details}

\label{sec:appendx_exp}
We evaluate our method upon Transformer-Base ~\citep{transformers} and conduct experiments under same hyper-parameters for fair comparison. We use fairseq~\citep{ott2019fairseq} as the main code base. 

Before training, we first apply BPE~\citep{bpe} to tokenize the corpus for 16k steps each language and then learn a joint dictionary. During training, the label smoothing parameter $\alpha$ is set to 0.1 except for Table~\ref{tab:1}'s exploration in alpha values. We use Adam optimizer with betas to be (0.9,0.98) and learning rate is 0.0007. During warming up steps, the initial learning rate is 1e-7 and there are 1000 warm-up steps. We use a batch-size of 2048 together with an update-freq of 4 on two NVIDIA 3090 GPUs. Dropout rate is set to 0.3 and weight decay is set to 0.0001 for all experiments. We average the last 3 checkpoints to generate the final model in the main bilingual experiments before inferring on the test set. We use beam size as 5 during all testing.

% This is an appendix.

\end{document}